\documentclass{article}
\usepackage{graphicx}
\usepackage{booktabs}
\usepackage{multirow}
\usepackage{caption}
\usepackage{adjustbox}
\usepackage{amsmath}
\usepackage{floatflt}
\usepackage{wrapfig}
\usepackage{hyperref}
\usepackage{url}
\newcommand{\nop}[1]{}

\PassOptionsToPackage{numbers, compress}{natbib}
\usepackage[preprint]{neurips_2024}

\title{Direct Multi-Token Decoding}

\author{Xuan Luo, Weizhi Wang, Xifeng Yan \\
Department of Computer Science, UC Santa Barbara \\
\{xuan\_luo, weizhiwang, xyan\}@cs.ucsb.edu}

\begin{document}

\maketitle

\begin{abstract}
Decoder-only transformers have become the standard architecture for large language models (LLMs) due to their strong  performance. Recent studies suggest that, in pre-trained LLMs, early, middle, and late layers may serve distinct roles: Early layers focus on understanding the input context, middle layers handle task-specific processing, and late layers convert abstract representations into output tokens.  We hypothesize that once representations have been processed by the early and middle layers, the resulting hidden states may encapsulate sufficient information to support the generation of multiple tokens using only the late layers, eliminating the need to repeatedly traverse the early and middle layers. We refer to this inference paradigm as Direct Multi-Token Decoding (DMTD). Unlike speculative decoding, our method introduces no additional parameters, auxiliary routines, or post-generation verification. Despite being trained on a limited dataset, a fine-tuned DMTD Qwen3-4B model has already demonstrated promising results, achieving up to a 2× speedup with only minor performance loss. Moreover, as shown in our scaling analysis, its performance is expected to further improve with larger training datasets.

\end{abstract}

\section{Introduction}

Transformers are the default choice for building large language models (LLMs). The original Transformer \citep{transformer} employed an encoder-decoder structure for sequence-to-sequence modeling, where the encoder processed input sequences for natural language understanding (NLU) and the decoder produces outputs for natural language generation (NLG). In this setup, the context was encoded once and repeatedly attended to during decoding. Subsequently, decoder-only architectures~\citep{gpt,llama,deepseekr1} have become the mainstream due to their simplicity and better scaling with training  data. It  leverages masked self-attention to process sequences causally, enabling efficient parallel computation during training and supporting versatile multi-task processing through prompting.

\begin{figure}[t]
\centering
\includegraphics[width=0.7\linewidth]{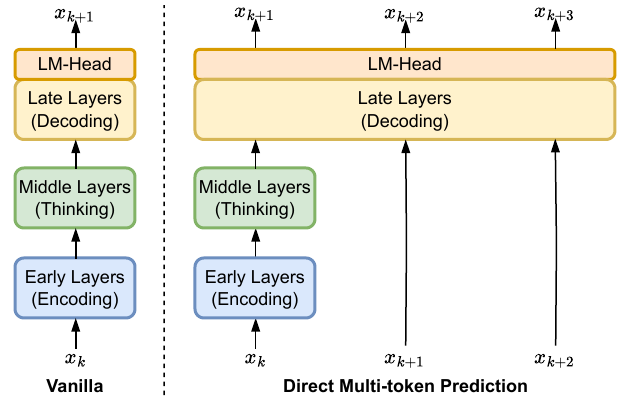}
\caption{Vanilla next token prediction vs. Direct Multi-Token Decoding.}
\label{fig:fig1}
\end{figure}

Recent studies reveal that decoder-only transformers may exhibit specialized functional roles across their layers~\citep{painter,primer,talkingheads,layerbylayer}. Specifically, these layers can be categorized into three functional stages. First, early layers encode syntactic and semantic features of the input context~\citep{early1,early2}. Next, middle layers handle reasoning and task-specific processing~\citep{flexidepth,middle}. Finally, late layers generate token-level predictions~\citep{late1,late2}. This layered specialization suggests that, while encoder-decoder architectures explicitly define encoding and decoding components, decoder-only models might implicitly develop a similar structure through training. To reflect their roles, we refer to these stages hypothetically as encoding, thinking, and decoding layers, respectively, as illustrated in Figure \ref{fig:fig1} (left), though there are no clear boundaries between these layers.

This implicit functional specialization also highlights potential inefficiencies in LLM's layer utilization. For instance, methods like FlexiDepth~\citep{flexidepth} have demonstrated that many layers can be dynamically skipped without significantly degrading performance. While LLMs utilize nearly all layers generating tokens that require complex computation, they can skip a substantial number of middle layers for simpler tasks like string copy. This finding aligns with intuition, as the difficulty to generate different tokens inherently varies.\nop{For straightforward tasks like string copying that require minimal reasoning, it is natural to bypass the middle layers associated with the thinking stage.} It indicates that spare computational cycles exist within the transformer's pipeline. This phenomenon motivated us to wonder: Could such underutlization be repurposed — that is, to encapsulate more information about future tokens in the current hidden states, and then allow subsequent tokens to attend to them through the decoding layers only, where multiple tokens can be generated?

In this work, we propose Direct Multi-Token Decoding (DMTD), which reuses the late layers to directly decode multiple tokens. Unlike the vanilla decoder-only transformer that generates tokens one by one through full forward passes, the proposed DMTD operates in fixed multi-token cycles. Figure \ref{fig:fig1} (right) demonstrates the generation pipeline of DMTD in a single cycle. DMTD performs only one full forward pass at the beginning of the cycle and then reuses the later layers to decode multiple tokens consecutively. This cycle-based setting transforms the irregular computational redundancies observed in pre-trained LLMs into a fixed periodical pattern for efficient decoding. DMTD features a minimal design, introducing no extra layers~\citep{eagle,eagle2,eagle3, deepseek-v3}, LM heads~\citep{medusa}, or post-processing routines like speculative decoding~\citep{speculative,eagle}. As the pre-trained LLMs are fine-tuned or continually trained under the DMTD framework, this approach eliminates the need for external adapters required by methods such as FlexiDepth~\citep{flexidepth} for layer skipping.  After training, it simply uses the tuned  neural network from the original model for multi-token decoding.

To support direct multi-token decoding, we trained the proposed DMTD in an end-to-end manner with all parameters tunable. The training initializes parameters from a pre-trained LLM and then fine-tunes on approximately 1.5B tokens. We found that through fine-tuning, our method can support sustained multi-token prediction with minor performance degradation. Furthermore, our scaling experiments demonstrate that the performance of our method improves continuously with increasing training data. As such, performing large-scale continued pre-training followed by post-training methods~\citep{deepseekr1,instructgpt} would be an effective approach to fully exploit the potential of this architecture.

We evaluated the proposed direct multi-token decoding across various benchmarks. By reusing the last 8 layers of the 36-layer Qwen3-4B~\citep{qwen3} and decoding two tokens per cycle, DMTD maintains 100\% of the original performance relative to the vanilla model. This performance retention remains strong at 98.4\% for three-token decoding cycles and 96.3\% for four-token decoding cycles. This cycled decoding approach reduces the total number of layers traversed during forward passes, enabling up to a 2× speedup in inference time with a cycle length of 4. We further observe that DMTD demonstrates relatively better performance on larger language models, suggesting promising directions for future exploration of its scalability on even larger architectures. To facilitate further research, we open-source our models and code at~\url{https://github.com/luoxuan-cs/Direct-Multitoken-Decoding}.

\section{Method}

In this section, we present the training and inference processes of the proposed direct multi-token decoding (DMTD).  During training, we use a cyclical masking strategy to  enable efficient learning of multiple future tokens. During inference, decoding proceeds sequentially across cycles, incorporating a cyclical refilling mechanism to recover missing KV cache entries, thereby supporting sustained generation without speculative decoding.

\subsection{Parallel Training with Cyclical Masking}
We propose a cyclical masking strategy to unify multi-token predictions within a single sequence during training. In standard next-token prediction, models learn to forecast one token at a time based on the preceding sequence. For multi-token prediction, our approach extends this by enabling the model to learn multiple future tokens simultaneously, all from the same input sequence. It does so by masking specific parts of the sequence intentionally, which directs the model to focus on predicting different future positions without needing separate sequences~\citep{eagle,medusa,deepseek-v3}. We define the cycle length of multi-token decoding as $\tau$. Figure~\ref{fig:fig2} illustrates the training pipeline for $\tau=3$. Given an input sequence $\mathbf{x} = {x_0, x_1, \ldots, x_n}$ and cycle length $\tau$, the training process consists of three phases:

\begin{wrapfigure}{r}{0.4\textwidth}
    \vspace{-0.0em}
    \centering
    \includegraphics[width=\linewidth]{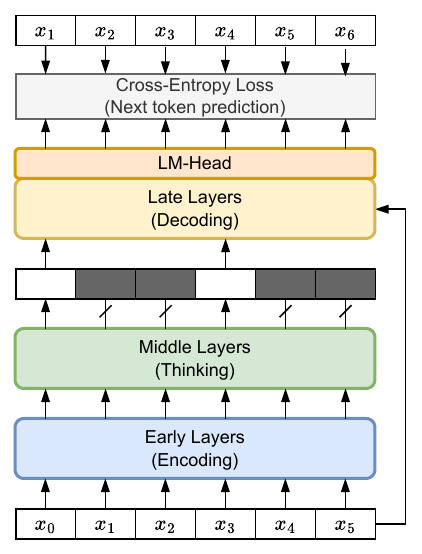}
    \vspace{0em}
    \caption{DMTD training pipeline with a cycle length of 3. The method requires no additional parameters and uses a single forward pass with masking to enable multi-token prediction training.}
    \vspace{0em}
    \label{fig:fig2}
\end{wrapfigure}

\textbf{Encoding Layers:} We first obtain the initial token embeddings from the input sequence using the embedding layer: $\mathbf{h}_{\text{emb}} = \text{Embed}(\mathbf{x})$. These embeddings are then processed by the encoding layers to produce the encoding representations $\mathbf{h}_{\text{enc}}$:

\begin{equation}
\mathbf{h}_{\text{enc}} = \text{EncodingLayers}(\mathbf{h}_{\text{emb}}).
\end{equation}

\textbf{Thinking Layers:} The encoding representations $\mathbf{h}_{\text{enc}}$ are further refined through the thinking layers to generate thinking representations $\mathbf{h}_{\text{think}}$:
\begin{equation}
\mathbf{h}_{\text{think}} = \text{ThinkingLayers}(\mathbf{h}_{\text{enc}}).
\end{equation}

\textbf{Decoding Layers with Masking:} Our training approach uses a masking strategy to simulate different execution paths within a single forward pass. A mask is applied based on the cycle length $\tau$ to selectively combine the input embeddings $\mathbf{h}_{\text{emb}}$ and thinking representations $\mathbf{h}_{\text{think}}$. For position indices $\mathbf{p} = \{0, 1, \ldots, n-1\}$, we create a binary pattern mask $\mathbf{M}$ where:
\begin{equation}
m_i = \begin{cases}
1 & \text{if } p_i \bmod \tau = 0, \\
0 & \text{otherwise}.
\end{cases}
\end{equation}

For example, with a cycle length of $\tau=3$, the masking pattern becomes $[1,0,0,1,0,0,1,0,0,\ldots]$. The masked hidden states are then computed as follows:
\begin{equation}
\mathbf{h}_{\text{masked}} = \mathbf{h}_{\text{emb}} + \mathbf{h}_{\text{think}} \odot \mathbf{M}.
\end{equation}

Alternatively, we can also leverage the encoding representations $\mathbf{h}_{\text{enc}}$ for multi-token decoding by using $\mathbf{h}_{\text{enc}}$ instead of $\mathbf{h}_{\text{emb}}$ to compute $\mathbf{h}_{\text{masked}}$. Under this setting, we will reuse the encoding layers as well as the decoding layers for multi-token decoding.

The resulting masked hidden states $\mathbf{h}_{\text{masked}}$ are then processed through the decoding layers and the LM head to obtain the output logits $\mathbf{z}$:
\begin{equation}
\mathbf{z} = \text{LMHead}(\text{DecodingLayers}(\mathbf{h}_{\text{masked}})).
\end{equation}

Finally, we will simply use the vanilla next token prediction loss for optimization:
\begin{equation}
\mathcal{L} = \frac{1}{n} \sum_{i=0}^{n-1} \text{CrossEntropy}(\mathbf{z}_i, x_{i+1}).
\end{equation}

Although we use the vanilla next-token prediction loss, the masking strategy enables the model to learn predictions for multiple future tokens. This paradigm differs from earlier approaches~\citep{mtp,medusa,deepseek-v3}, which rely on multiple divergent execution paths for multi-token prediction. In those methods, separate sequences and cross-entropy losses are necessary for optimization, leading to high GPU memory usage due to storing and processing multiple sequences. Our method unifies these paths into one sequence through cyclical masking and hidden state reuse, eliminating redundant computations by reusing shared prefix representations across all prediction levels.

\subsection{Multi-Token Decoding with cyclical refilling}

Direct multi-token decoding aims to avoid the step of post-generation verification required by speculative decoding~\citep{speculative,eagle}. It performs decoding in fixed multi-token cycles, leveraging the specialized roles of the late layers to generate multiple tokens efficiently. In each cycle, the first forward pass processes the input through all layers, while subsequent forward passes within the cycle use only the late layers. However, as generation progresses, skipping the early and middle layers results in missing entries in the key-value cache (KV-cache)~\citep{kvcache} for the early and middle layers, which stores intermediate representations the attention module needs for subsequent generation. These missing KV cache entries can degrade the quality of new tokens due to incomplete context. To address this, we introduce a cyclical refilling strategy that restores missing KV cache entries from previous cycles.

\begin{figure}[t]
\centering
\includegraphics[width=0.8\linewidth]{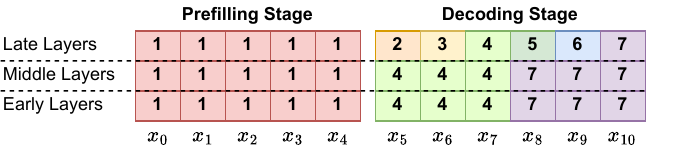}
\caption{Cyclical refilling for multi-token decoding with a cycle length of 3. There are three blocks within each column, representing the early, middle, and late layers. Blocks with the same color are computed in the same forward pass. The numbers on the blocks represent the index of the forward pass. }
\label{fig:fig3}
\end{figure}

Figure \ref{fig:fig3} outlines the decoding process of our method. Consider an input context of tokens \( x_0, x_1, x_2, x_3, x_4 \). The process begins by forwarding all the input context through the early, middle, and late layers for prefilling, which also generates the first output token \( x_5 \) and serves as the initial forward pass of the first generation cycle. In the subsequent decoding stage of this cycle, the model forwards \( x_5 \) through only the late layers to produce \( x_6 \), and then forwards \( x_6 \) through the late layers to generate \( x_7 \). This completes the first cycle, which involves three forward passes in total. At the start of the second cycle, our model forwards \( x_5, x_6, x_7 \) together through the early and middle layers to refill the KV cache for \( x_5 \) and \( x_6 \), while only \( x_7 \) is processed through the late layers to generate \( x_8 \). Similarly, \( x_8 \) and \( x_9 \) are forwarded through only the late layers in the second cycle, with their KV cache refilled in the subsequent cycle. This refilling mechanism ensures that the full context remains available for generation, eliminating the need for speculative decoding methods to mitigate error propagation.

We can find that the overall computational load of the proposed DMTD is the same as the vanilla transformer. Therefore, how can our method achieve faster inference? The reason lies in the memory-bound nature of large language model (LLM) inference~\citep{speculative,eagle,medusa}. In this scenario, GPU computational resources are underutilized, and inference speed depends primarily on the number of modules processed rather than the total computational volume~\citep{mind}. For instance, on modern GPUs, forwarding three tokens through 32 transformer layers takes roughly the same time as forwarding one token through the 32 layers. In contrast, forwarding one token through 64 layers requires approximately twice the time as forwarding one through 32 layers~\citep{kvcache,trainlarge}. This memory-bound characteristic also explains why speculative decoding achieves speedup despite additional computations for drafting and verification~\citep{speculative}. By processing fewer layers per token, DMTD capitalizes on this property to accelerate inference, even with a computational load equivalent to that of a vanilla transformer.

To quantify the efficiency of DMTD, we introduce the concept of Percentage of Layers per Token (PLT), which measures the average number of transformer layers processed per generated token. A lower PLT indicates higher efficiency in a memory-bound scenario. Let \( L \) denote the total number of transformer layers, with \( L_e \), \( L_t \), and \( L_d \) representing the number of encoding, thinking, and decoding layers, respectively. In a vanilla decoder-only transformer, each token is processed through all layers, resulting in an average of \( L_e + L_t + L_d = L \) layers per token, yielding a PLT of 1. In DMTD, token generation occurs in cycles, each containing $\tau$ tokens. Within each cycle, the first token is processed through all $L$ layers, while the remaining $\tau - 1$ tokens are processed only through the $L_d$ decoding layers. Thus, the PLT for DMTD is:

\begin{equation}
PLT = \frac{L+(\tau-1)L_d}{\tau L} = \frac{1}{\tau} + \frac{\tau - 1}{\tau} \cdot \frac{L_d}{L}.
\label{eq:eq7}
\end{equation}

This expression shows that the PLT of DMTD depends on the cycle length \( \tau \) and the ratio \( \frac{L_d}{L} \). A larger cycle length $\tau$ or a smaller proportion of decoding layers \( \frac{L_d}{L} \) reduces the PLT, thereby enhancing the inference efficiency of DMTD.  More speedup will likely be expected if there are more layers in a pre-trained LLM.

\section{Experiments}
\subsection{Implementation Details}
We implement DMTD on the pre-trained Qwen3-4B model~\citep{qwen3}, which consists of 36 transformer layers. To enable multi-token decoding, we reuse the latter 8 layers as decoding layers, as prior works~\citep{painter,layerbylayer} and empirical studies suggest that late layers are specialized for token-level predictions. The default cycle length is set to 3, allowing each cycle to generate 3 tokens, as illustrated in Figures~\ref{fig:fig2} and~\ref{fig:fig3}. We train DMTD using supervised fine-tuning (SFT) on the AM-Thinking-v1-Distilled dataset~\citep{am} for 1 epoch using the AdamW~\citep{adamw} optimizer with a learning rate of $1 \times 10^{-4}$, max gradient norm of 1.0, $\beta_1 = 0.9$, $\beta_2 = 0.95$. We use a warmup ratio of 0.1, a cosine learning rate scheduler, and a global batch size of 512.

We assess our method on ARC-Easy, ARC-Challenge~\citep{arc}, WinoGrande~\citep{winogrande}, GSM8K~\citep{gsm8k}, and CoQA~\citep{coqa}. ARC~\citep{arc} examines knowledge and reasoning through grade-school science questions. WinoGrande~\citep{winogrande} tests commonsense reasoning with adversarial Winograd schema challenges. GSM8K~\citep{gsm8k} evaluates multi-step mathematical reasoning using grade-school word problems. CoQA~\citep{coqa} measures conversational question-answering skills, including coreference and pragmatic reasoning. We apply 4-shot prompting for GSM8K, while the others are evaluated in a zero-shot setting. All the evaluations are conducted under a batch size of 32. We utilize the think mode~\citep{qwen3} with chain-of-thought prompt~\citep{cot} for all benchmarks except CoQA, which does not require deep reasoning. All tasks involve continuous generation to assess the multi-token decoding capability of the proposed method.

\subsection{Decoding Cycle Length}
We evaluate the performance of DMTD across various cycle lengths using a default setup with 8 decoding layers. Models are trained and tested with cycle lengths of 2, 3, 4, and 6, denoted as MTD2, MTD3, MTD4, and MTD6, respectively. Table~\ref{tab:cycle_lengths} shows the results, with each model using a consistent cycle length for both training and evaluation. All generations are performed directly without post-verification. The overall score reflects the average relative performance compared to the vanilla Qwen3-4B~\citep{qwen3}.

As shown in Table~\ref{tab:cycle_lengths}, our proposed method performs effectively for cycle lengths up to 4, with performance gradually declining as the cycle length increases, maintaining 96.3\% of the vanilla model's overall performance at a cycle length of 4. However, performance noticeably drops beyond this point, falling to 82.1\% at a cycle length of 6. We hypothesize that this decline results from the limited dimensionality of the hidden states, which restricts their capacity to capture sufficient information about future tokens, thus hindering effective long-range multi-token generation. Full-scale pre-training of the proposed method on larger models could potentially support longer prediction horizons.

\begin{table}[t]
\centering
\caption{Performance across benchmarks for different cycle lengths.}
\label{tab:cycle_lengths}
\begin{adjustbox}{max width=\textwidth}
\begin{tabular}{lcccccc}
\toprule
 & ARC-E & ARC-C & WinoGrande & GSM8K & CoQA & Overall \\
\midrule
Vanilla  & 0.934 & 0.922 & 0.657 & 0.907 & 0.805 & 100\% \\
MTD2     & 0.930 & 0.897 & 0.701 & 0.901 & 0.798 & 100.0\% \\
MTD3     & 0.921 & 0.886 & 0.673 & 0.889 & 0.780 & 98.4\% \\
MTD4     & 0.916 & 0.881 & 0.652 & 0.866 & 0.749 & 96.3\% \\
MTD6     & 0.872 & 0.801 & 0.601 & 0.500 & 0.672 & 82.1\% \\
\bottomrule
\end{tabular}
\end{adjustbox}
\end{table}

\subsection{Impact of Encoding and Decoding Layer Allocation}

In this section, we examine the effects of varying the allocation of encoding and decoding layers in the proposed method. This analysis aims to elucidate the relative importance of early (encoding) and late (decoding) layers in facilitating multi-token generation. We evaluate three configurations based on the total number of layers reused for tokens beyond the first: 4 layers, 8 layers, and 16 layers. The cycle length is fixed at 3 for all experiments. We denote configurations as E$x$D$y$, where $x$ represents the number of encoding layers and $y$ the number of decoding layers reused. Table~\ref{tab:layer_allocation} presents the performance across the benchmarks.

\begin{table}[t]
\centering
\caption{Performance comparison with different allocations of encoding and decoding layers.}
\begin{adjustbox}{max width=\textwidth}
\begin{tabular}{lcccccc}
\toprule
 & ARC-E & ARC-C & WinoGrande & GSM8K & CoQA & Overall \\
\midrule
Vanilla  & 0.934 & 0.922 & 0.657 & 0.907 & 0.805 & 100\% \\
\midrule
\multicolumn{7}{c}{\textbf{Reuse 4 Layers}} \\
\midrule
E4D0 & 0.412 & 0.364 & 0.517 & 0.048 & 0.532 & 46.7\% \\
E2D2 & 0.919 & 0.878 & 0.663 & 0.878 & 0.793 & 98.0\% \\
E0D4 & 0.918 & 0.882 & 0.665 & 0.889 & 0.758 & 97.5\% \\
\midrule
\cmidrule{1-7}
\multicolumn{7}{c}{\textbf{Reuse 8 Layers}} \\
\midrule
E8D0 & 0.540 & 0.470 & 0.497 & 0.194 & 0.604 & 56.2\% \\
E4D4 & 0.922 & 0.876 & 0.670 & 0.890 & 0.808 & 98.8\% \\
E0D8 & 0.921 & 0.886 & 0.673 & 0.889 & 0.780 & 98.4\% \\
\midrule
\cmidrule{1-7}
\multicolumn{7}{c}{\textbf{Reuse 16 Layers}} \\
\midrule
E16D0 & 0.741 & 0.609 & 0.535 & 0.544 & 0.717 & 75.2\% \\
E8D8 & 0.921 & 0.890 & 0.683 & 0.898 & 0.812 & 99.8\% \\
E0D16 & 0.928 & 0.898 & 0.685 & 0.907 & 0.802 & 100.1\%  \\
\bottomrule
\end{tabular}
\end{adjustbox}
\label{tab:layer_allocation}
\end{table}

The results demonstrate that at least a few decoding layers are necessary for effective multi-token decoding. Reusing only the encoding layers (e.g., E$x$D0 configurations) yields suboptimal performance, reaching only 75.2\% even with 16 encoding layers (E16D0). It indicates that, once the input context is processed, scaling only the early layers is insufficient for accurate token prediction. In contrast, configurations that emphasize decoding layers (e.g., E0D$y$) maintain performance close to the vanilla baseline. Scaling both encoding and decoding layers produces similar outcomes to reusing primarily decoding layers; for instance, E8D8 achieves 99.8\% overall performance, comparable to E0D16's 100.1\%.

\begin{figure}[t]
\centering
\includegraphics[width=\linewidth]{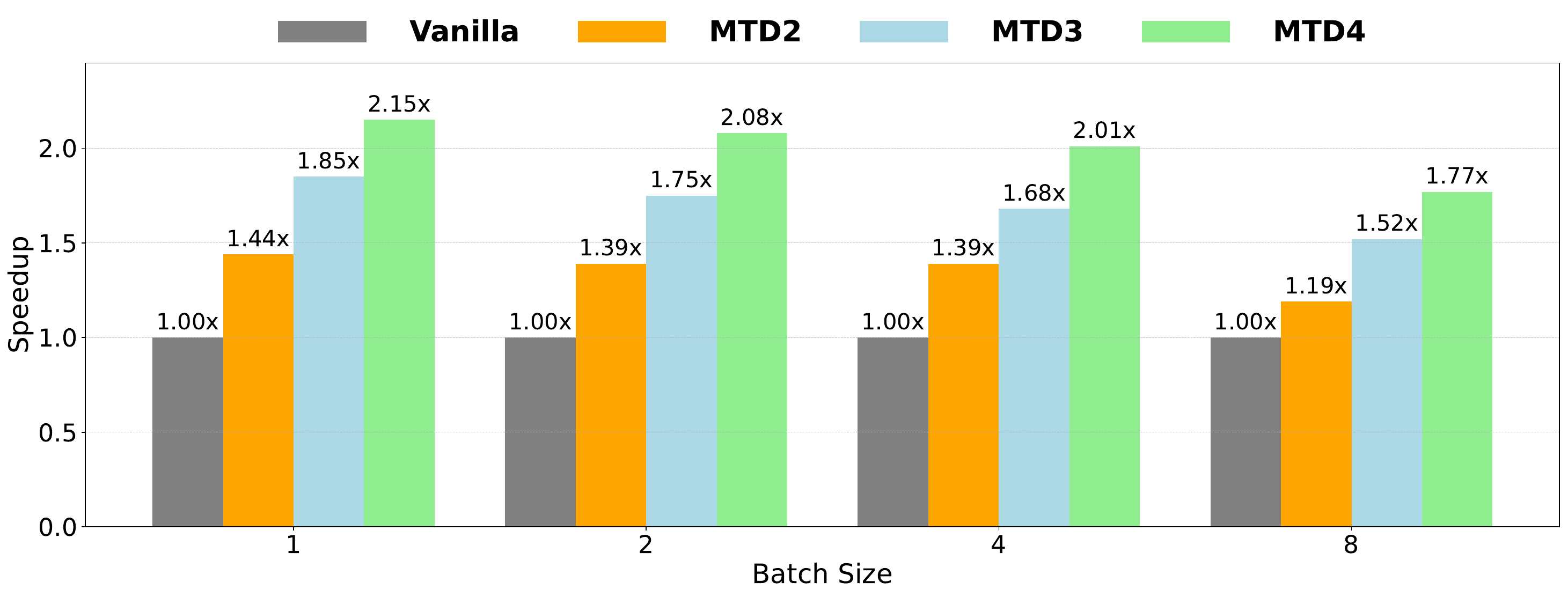}
\caption{Speedup comparison.}
\label{fig:fig4}
\end{figure}

\subsection{Inference Speedup}
To evaluate the inference efficiency, we compare the default configuration that reuses the last 8 layers against the vanilla Qwen3-4B model. All evaluations are conducted on a single NVIDIA A100-40GB GPU, using a static input length of 1024 tokens (randomly sampled from the vocabulary) and a generation length of 1024 tokens. We evaluated the throughput of the proposed method with different cycle length. Table~\ref{tab:throughput} reports the throughput for these models.

\begin{table}[ht]
\centering
\caption{Throughput (tokens per second) comparison of our method and Qwen3-4B.}
\label{tab:throughput}
\begin{adjustbox}{max width=\textwidth}
\begin{tabular}{lcccc}
\toprule
\ & Batch=1 & Batch=2 & Batch=4 & Batch=8 \\
\midrule
Vanilla & 21.83 & 44.69 & 90.76 & 181.03 \\
MTD2 & 31.47 & 61.95 & 126.04 & 214.54 \\
MTD3 & 40.49 & 78.29 & 152.59 & 275.52 \\
MTD4 & 47.04 & 92.75 & 183.16 & 320.12 \\
\bottomrule
\end{tabular}
\end{adjustbox}
\end{table}

As shown in Table~\ref{tab:throughput} and Figure~\ref{fig:fig4}, our method achieves notable speedups, with improvements increasing with cycle length, particularly at lower batch sizes. For example, MTD4 provides up to 2.15$\times$ speedup at batch size 1. At lower batch sizes, these gains align with the theoretical speedup based on the Percentage of Layers per Token (PLT) in memory-bound regimes—for instance, for MTD3, the PLT is approximately 0.48, with its inverse of 2.08 aligning with the observed 1.85$\times$ speedup. As batch size increases, the system becomes more compute-bound, leading to reduced relative gains. For example, for MTD4, the speedup gradually drops from 2.15$\times$ at batch size 1 to 1.77$\times$ at batch size 8 compared to the vanilla model.

\subsection{Scaling with Training Data}
In this section, we investigate the scaling behavior of the proposed method as the volume of training data increases. We conduct experiments using the E0D8MTD3 configuration across three model sizes: Qwen3-0.6B, Qwen3-1.7B, and Qwen3-4B. Our hypothesis is that larger training datasets will lead to improved model performance, as indicated by reductions in cross-entropy loss. Figure~\ref{fig:fig4} depicts these scaling curves for the different model sizes.

The results reveal a consistent decrease in cross-entropy loss as training data increases for all model sizes, with the trends approximating log-linear relationships. To quantify the goodness of fit, we perform linear regression on each curve and report the slope, indicating the rate of loss reduction per order of magnitude increase in tokens, and the coefficient of determination $R^2$, which measures how well the linear model explains the observed loss variations, with values closer to 1 indicating a strong fit. For the 0.6B model, the slope is -0.179 with an $R^2$ of 0.966; for the 1.7B model, the slope is -0.191 with an $R^2$ of 0.972; and for the 4B model, the slope is -0.178 with an $R^2$ of 0.994. The high $R^2$ values, particularly exceeding 0.96 across all models, suggest that the loss reduction follows a highly predictable pattern as training data scales. Given that our current experiments are conducted with supervised fine-tuning on a limited dataset due to resource constraints, we expect that access to larger-scale continued pre-training, followed by post-training alignment, would further enhance these trends, potentially unlocking greater multi-token prediction capabilities on larger models.

\begin{figure}[t]
\centering
\includegraphics[width=0.6\linewidth]{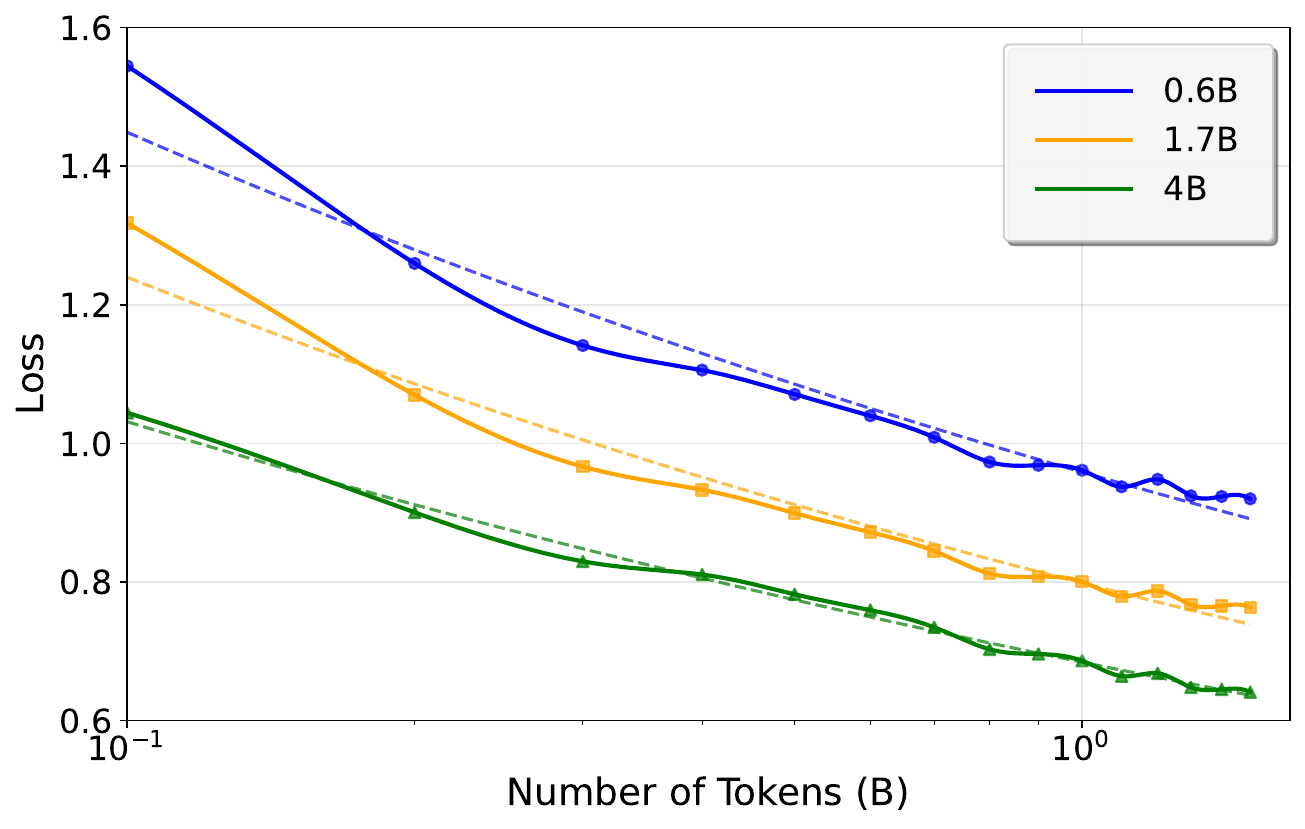}
\caption{Scaling law of the proposed Direct Multi-token Decoding. The x-axis represents the number of training tokens (in billions) on a logarithmic scale, while the y-axis shows the cross-entropy loss.}
\label{fig:fig5}
\end{figure}

\subsection{Impact of Model Scale on Direct Multi-Token Decoding}
In this section, we evaluate the performance of the proposed method with the default E0D8 configuration and a cycle length of 3 (E0D8MTD3) across language models of varying sizes: Qwen3-0.6B, Qwen3-1.7B, and Qwen3-4B. This setup enables us to assess how the effectiveness of direct multi-token decoding scales with model size, particularly in terms of maintaining performance across diverse benchmarks.

\begin{table}[t]
\centering
\caption{Performance of E0D8MTD3 across different Qwen3 model sizes.}
\label{tab:model_scale}
\begin{adjustbox}{max width=\textwidth}
\begin{tabular}{lcccccc}
\toprule
 & ARC-E & ARC-C & WinoGrande & GSM8K & CoQA & Overall \\
\midrule
\multicolumn{7}{c}{\textbf{Qwen3-0.6B}} \\
\midrule
Vanilla & 0.813 & 0.687 & 0.499 & 0.751 & 0.706 & 100\% \\
Ours & 0.687 & 0.562 & 0.489 & 0.532 & 0.550 & 72.6\% \\
\midrule
\multicolumn{7}{c}{\textbf{Qwen3-1.7B}} \\
\midrule
Vanilla & 0.910 & 0.852 & 0.566 & 0.828 & 0.776 & 100\% \\
Ours & 0.855 & 0.773 & 0.540 & 0.716 & 0.715 & 91.7\% \\
\midrule
\multicolumn{7}{c}{\textbf{Qwen3-4B}} \\
\midrule
Vanilla & 0.934 & 0.922 & 0.657 & 0.907 & 0.805 & 100\% \\
Ours & 0.921 & 0.886 & 0.673 & 0.889 & 0.780 & 98.4\% \\
\bottomrule
\end{tabular}
\end{adjustbox}
\end{table}

From the results in Table~\ref{tab:model_scale}, we can clearly observe that, under the same configuration, larger models benefit more from our method. For instance, the Qwen3-4B model retains 98.4\% of the vanilla performance, compared to only 72.6\% for the Qwen3-0.6B model. Even though the Qwen3-0.6B and Qwen3-1.7B models have only 28 transformer layers, where reusing 8 decoding layers constitutes a larger proportion of the total architecture (approximately 28.6\%), they exhibit worse relative performance than the Qwen3-4B model with its 36 layers, where 8 layers represent about 22.2\%. We hypothesize that this is due to the increased number of parameters and larger dimensionality in bigger models, which allow them to encode richer anticipatory information, thereby better supporting multi-token prediction. Additionally, for transformers with more layers, reusing the same fixed number of decoding layers results in a lower Percentage of Layers per Token (PLT) as defined in Equation~\ref{eq:eq7}, leading to higher potential speedups. These results indicate that our method is particularly well-suited for larger LLMs, and experiments on even bigger models may yield further improvements.

\subsection{Inference Cycle Length}

In this section, we investigate how DMTD will perform when the inference cycle length differs from the training cycle length. Specifically, we train the model with the default E0D8MTD3 setting. During inference, we evaluate its performance across different inference cycle lengths to assess its robustness. Table~\ref{tab:generalization_cycle_lengths} presents the results.

\begin{table}[ht]
\centering
\caption{Performance across benchmarks for different inference cycle lengths. The notation $k \to z$ denotes training on a cycle length of $k$ and evaluation on a cycle length of $z$.}
\label{tab:generalization_cycle_lengths}
\begin{adjustbox}{max width=\textwidth}
\begin{tabular}{lcccccc}
\toprule
 & ARC-E & ARC-C & WinoGrande & GSM8K & CoQA & Overall \\
\midrule
Vanilla & 0.934 & 0.922 & 0.657 & 0.907 & 0.805 & 100\% \\
3 \(\to\) 2 & 0.926 & 0.893 & 0.695 & 0.912 & 0.794 & 100.1\% \\
3 \(\to\) 3 & 0.921 & 0.886 & 0.673 & 0.889 & 0.780 & 98.4\% \\
3 \(\to\) 4 & 0.910 & 0.863 & 0.620 & 0.759 & 0.738 & 92.2\% \\
3 \(\to\) 5 & 0.867 & 0.809 & 0.572 & 0.400 & 0.716 & 80.1\% \\
3 \(\to\) 6 & 0.797 & 0.717 & 0.512 & 0.149 & 0.696 & 68.8\% \\
\bottomrule
\end{tabular}
\end{adjustbox}
\end{table}

From Table~\ref{tab:generalization_cycle_lengths}, we observe that the model trained with a cycle length of 3 can generalize effectively to both shorter and longer cycle lengths, albeit with varying degrees of performance retention. At the trained length of 3, it retains 98.4\% performance, while extending to a cycle length of 4 yields 92.2\%—a modest degradation that still preserves strong capabilities across most benchmarks. However, further extension to lengths of 5, 6 results in sharper declines, particularly evident in reasoning-intensive tasks like GSM8K. This suggests that the thinking layers, when trained to encode anticipatory information for a cycle length of 3, possess sufficient flexibility to support multi-token prediction across neighboring inference cycle lengths.  Interestingly, this flexibility allows a single model to dynamically adjust the inference cycle length and achieve the desired balance between speedup and quality.

\section{Related Works}

\subsection{Large Language Models}
Initially, the Transformer architecture~\citep{transformer} adopted an encoder-decoder structure for sequence-to-sequence modeling, where the encoder uses bidirectional attention to understand the input context, and the decoder applies causal attention for token generation. However, the vanilla transformer supports only single-task learning, requiring one model per task. Building upon this, T5~\citep{t5} employs a unified text-to-text framework to handle multiple tasks within a single model using task-specific prefixes. Subsequently, FLAN~\citep{flan} introduced the concept of instruction tuning by fine-tuning models on diverse tasks with natural language instructions, enhancing zero-shot and few-shot performance. Currently, the encoder-decoder architecture remains widely used in multimodal language models~\citep{blip,blip2,flamingo,valm} and latency-sensitive applications~\citep{edgemma}. 

In contrast, decoder-only models, such as the GPT series~\citep{gpt,instructgpt}, emerged to prioritize generative tasks by relying solely on causal self-attention, enabling scalable, prompt-driven learning with reduced architectural complexity. These models excel in open-ended tasks like dialogue and text generation, dominating modern LLM applications~\citep{llama,gemma,qwen}. Recent studies reveal that decoder-only models implicitly develop a three-layer functional specialization during training, mirroring an encoding-thinking-decoding pipeline. Early layers focus on syntactic and semantic encoding, transforming raw inputs into a stable embedding space critical for contextual understanding~\citep{painter}. Middle layers handle reasoning and task-specific abstraction, compressing information and enabling complex processing, such as multi-step reasoning, with greater robustness to layer manipulation~\citep{layerbylayer}. Late layers specialize in token-level predictions, refining representations for generation but often discarding broader contextual features~\citep{painter,layerbylayer}. Mixture-of-Depths (MoD) demonstrates that, with large-scale pre-training, we can use a router to skip the redundant layers in transformers. Furthermore, FlexiDepth~\citep{flexidepth} demonstrates that these layers exhibit varying sparsity in a bowl-like pattern. However, they represent irregular skipping patterns, which are difficult to provide acceleration in memory-bound scenarios. These findings inspired our Direct Multi-Token Decoding (DMTD) method, which leverages this layer specialization by cyclically reusing late layers to efficiently generate multiple tokens, repurposing underutilized computations in pre-trained LLMs without additional components.

\subsection{Multi-token Prediction}

To accelerate inference in decoder-only models, recent works have explored multi-token prediction techniques that enable parallel generation of multiple tokens, addressing the memory bottlenecks of autoregressive decoding~\citep{speculative}. Early approaches, such as speculative decoding~\citep{speculative}, leverage smaller draft models to propose multiple candidate tokens in parallel, verifying them against the target LLM to achieve speedups without altering output distributions. Building on this, methods like Medusa~\citep{medusa} introduce multiple decoding heads on top of the LLM to predict several subsequent tokens simultaneously, using tree-based attention for verification and fine-tuning strategies to balance accuracy and efficiency. Similarly, the EAGLE seires~\citep{eagle,eagle2,eagle3} rethinks speculative sampling at the feature level, resolving uncertainty in intermediate representations by advancing token sequences.  A comprehensive examination of efficient speculative decoding for models like Llama at scale was given  by~\citep{llamaeagle}.

In parallel, training-focused innovations incorporate multi-token prediction as an auxiliary objective, as in DeepSeek-V3~\citep{deepseek-v3}, where it enhances performance in large MoE models by predicting multiple future tokens during pre-training, or as a core loss in models trained to forecast future tokens via independent heads~\citep{mtp}. These techniques collectively demonstrate substantial inference speedups and improved generative capabilities, inspiring our DMTD to directly reuse existing late layers for cyclical multi-token decoding, avoiding the need for auxiliary models or heads while exploiting underutilized layers.

\section{Limitations}

Since the pre-training dataset of Qwen3 is not publicly available (Qwen3 utilizes a large-scale training dataset consisting of approximately 36 trillion
tokens. We only used 1.5B tokens), we were unable to conduct full continual training on the complete dataset to assess the performance of a fully developed DMTD. As shown in Figure \ref{fig:fig5}, its performance is expected to further improve with access to additional training data. For this reason, we do not provide a direct experimental comparison between our method and speculative decoding, leaving such an evaluation to future work. Nevertheless, if required, speculative decoding can still be applied to the tokens generated by DMTD.\nop{Under the current experimental setup, DMKD underperforms state-of-the-art speculative decoding methods for small batch-size.}  Overall, DMKD offers a simple paradigm that merits further investigation, especially in the context of MoE of experts and large batch size, where the performance of speculative decoding might decrease with increasing batch size \citep{eagle3, llamaeagle}.

\nop{
\section{Discussion}
Introduce the concept of using the last layer and the embedding of token to better extract the meaning of this token in the existing context. 
}

\section{Conclusion}
In this work, we introduce Direct Multi-Token Decoding (DMTD), a paradigm that enables sustained multi-token generation without introducing additional parameters or requiring post-generation verification, as is the case with speculative decoding. Instead, DMTD leverages the inherent underutilization present in pre-trained LLMs and repurposes it into fixed cycles of multi-token generation. Our experimental results demonstrate not only the feasibility of this approach but also suggest that its performance can further improve with larger training datasets, opening a promising new direction for accelerating LLM inference.

\section*{Acknowledgements}
We would like to thank Facebook (now Meta) for donating the A100-40G GPUs used in our experiments. We also gratefully acknowledge the generous support of the NVIDIA Academic Grant Program. Access to NVIDIA GPUs and software toolkits enabled us to conduct experiments on larger training datasets and inspired new research directions.

\newpage
\bibliography{references}

\begin{thebibliography}{10}

\bibitem{flamingo}
Jean-Baptiste Alayrac, Jeff Donahue, Pauline Luc, Antoine Miech, and others.
\newblock Flamingo: a visual language model for few-shot learning.
\newblock In {\em Advances in Neural Information Processing Systems}, 2022.

\bibitem{early2}
Amos Azaria and Tom Mitchell.
\newblock The internal state of an {LLM} knows when it's lying.
\newblock In {\em The 2023 Conference on Empirical Methods in Natural Language Processing}, 2023.

\bibitem{qwen}
Jinze Bai, Shuai Bai, Yunfei Chu, Zeyu Cui, Kai Dang, Xiaodong Deng, Yang Fan, Wenbin Ge, Yu~Han, Fei Huang, et~al.
\newblock Qwen technical report.
\newblock {\em arXiv preprint arXiv:2309.16609}, 2023.

\bibitem{gpt}
Tom Brown, Benjamin Mann, Nick Ryder, Melanie Subbiah, Kaplan, et~al.
\newblock Language models are few-shot learners.
\newblock {\em Advances in neural information processing systems}, 2020.

\bibitem{medusa}
Tianle Cai, Yuhong Li, Zhengyang Geng, Hongwu Peng, Jason~D. Lee, Deming Chen, and Tri Dao.
\newblock Medusa: Simple {LLM} inference acceleration framework with multiple decoding heads.
\newblock In {\em Forty-first International Conference on Machine Learning}, 2024.

\bibitem{late1}
Yung-Sung Chuang, Yujia Xie, Hongyin Luo, et~al.
\newblock Dola: Decoding by contrasting layers improves factuality in large language models.
\newblock In {\em The Twelfth International Conference on Learning Representations}, 2024.

\bibitem{arc}
Peter Clark, Isaac Cowhey, Oren Etzioni, Tushar Khot, Ashish Sabharwal, Carissa Schoenick, and Oyvind Tafjord.
\newblock Think you have solved question answering? try arc, the ai2 reasoning challenge.
\newblock {\em arXiv preprint arXiv:1803.05457}, 2018.

\bibitem{gsm8k}
Karl Cobbe, Vineet Kosaraju, Mohammad Bavarian, Mark Chen, Heewoo Jun, Lukasz Kaiser, Matthias Plappert, Jerry Tworek, Jacob Hilton, Reiichiro Nakano, et~al.
\newblock Training verifiers to solve math word problems.
\newblock {\em arXiv preprint arXiv:2110.14168}, 2021.

\bibitem{primer}
Javier Ferrando, Gabriele Sarti, Arianna Bisazza, and Marta~R. Costa{-}juss{\`{a}}.
\newblock A primer on the inner workings of transformer-based language models.
\newblock {\em arXiv preprint arXiv:2405.00208}, 2024.

\bibitem{mtp}
Fabian Gloeckle, Badr~Youbi Idrissi, Baptiste Roziere, David Lopez-Paz, and Gabriel Synnaeve.
\newblock Better \& faster large language models via multi-token prediction.
\newblock In {\em Forty-first International Conference on Machine Learning}, 2024.

\bibitem{deepseekr1}
Daya Guo, Dejian Yang, Haowei Zhang, Junxiao Song, Ruoyu Zhang, Runxin Xu, Qihao Zhu, Shirong Ma, Peiyi Wang, Xiao Bi, et~al.
\newblock Deepseek-r1: Incentivizing reasoning capability in llms via reinforcement learning.
\newblock {\em arXiv preprint arXiv:2501.12948}, 2025.

\bibitem{early1}
Xuming Hu, Junzhe Chen, Xiaochuan Li, Yufei Guo, Lijie Wen, Philip~S. Yu, and Zhijiang Guo.
\newblock Towards understanding factual knowledge of large language models.
\newblock In {\em The Twelfth International Conference on Learning Representations}, 2024.

\bibitem{speculative}
Yaniv Leviathan, Matan Kalman, and Yossi Matias.
\newblock Fast inference from transformers via speculative decoding.
\newblock In {\em International Conference on Machine Learning}, 2023.

\bibitem{blip2}
Junnan Li, Dongxu Li, Silvio Savarese, and Steven Hoi.
\newblock Blip-2: Bootstrapping language-image pre-training with frozen image encoders and large language models.
\newblock In {\em International conference on machine learning}, 2023.

\bibitem{blip}
Junnan Li, Dongxu Li, Caiming Xiong, and Steven Hoi.
\newblock Blip: Bootstrapping language-image pre-training for unified vision-language understanding and generation.
\newblock In {\em International conference on machine learning}, 2022.

\bibitem{eagle2}
Yuhui Li, Fangyun Wei, Chao Zhang, and Hongyang Zhang.
\newblock Eagle-2: Faster inference of language models with dynamic draft trees.
\newblock {\em arXiv preprint arXiv:2406.16858}, 2024.

\bibitem{eagle}
Yuhui Li, Fangyun Wei, Chao Zhang, and Hongyang Zhang.
\newblock {EAGLE}: Speculative sampling requires rethinking feature uncertainty.
\newblock In {\em Forty-first International Conference on Machine Learning}, 2024.

\bibitem{eagle3}
Yuhui Li, Fangyun Wei, Chao Zhang, and Hongyang Zhang.
\newblock Eagle-3: Scaling up inference acceleration of large language models via training-time test.
\newblock {\em arXiv preprint arXiv:2503.01840}, 2025.

\bibitem{trainlarge}
Zhuohan Li, Eric Wallace, Sheng Shen, Kevin Lin, Kurt Keutzer, Dan Klein, and Joey Gonzalez.
\newblock Train big, then compress: Rethinking model size for efficient training and inference of transformers.
\newblock In {\em International Conference on machine learning}, 2020.

\bibitem{deepseek-v3}
Aixin Liu, Bei Feng, Bing Xue, Bingxuan Wang, Bochao Wu, Chengda Lu, Chenggang Zhao, Chengqi Deng, Chenyu Zhang, Chong Ruan, et~al.
\newblock Deepseek-v3 technical report.
\newblock {\em arXiv preprint arXiv:2412.19437}, 2024.

\bibitem{adamw}
Ilya Loshchilov and Frank Hutter.
\newblock Decoupled weight decay regularization.
\newblock In {\em International Conference on Learning Representations}, 2019.

\bibitem{flexidepth}
Xuan Luo, Weizhi Wang, and Xifeng Yan.
\newblock Adaptive layer-skipping in pre-trained {LLM}s.
\newblock In {\em Second Conference on Language Modeling}, 2025.

\bibitem{talkingheads}
Jack Merullo, Carsten Eickhoff, and Ellie Pavlick.
\newblock Talking heads: Understanding inter-layer communication in transformer language models.
\newblock In {\em The Thirty-eighth Annual Conference on Neural Information Processing Systems}, 2024.

\bibitem{instructgpt}
Long Ouyang, Jeffrey Wu, Xu~Jiang, Diogo Almeida, Carroll Wainwright, Mishkin, et~al.
\newblock Training language models to follow instructions with human feedback.
\newblock {\em Advances in neural information processing systems}, 2022.

\bibitem{kvcache}
Reiner Pope, Sholto Douglas, Aakanksha Chowdhery, Jacob Devlin, James Bradbury, Jonathan Heek, Kefan Xiao, Shivani Agrawal, and Jeff Dean.
\newblock Efficiently scaling transformer inference.
\newblock {\em Proceedings of machine learning and systems}, 2023.

\bibitem{t5}
Colin Raffel, Noam Shazeer, Adam Roberts, Katherine Lee, Sharan Narang, Michael Matena, Yanqi Zhou, Wei Li, and Peter~J Liu.
\newblock Exploring the limits of transfer learning with a unified text-to-text transformer.
\newblock {\em Journal of machine learning research}, 2020.

\bibitem{mind}
Pol~G Recasens, Ferran Agullo, Yue Zhu, Chen Wang, Eun~Kyung Lee, Olivier Tardieu, Jordi Torres, and Josep~Ll Berral.
\newblock Mind the memory gap: Unveiling gpu bottlenecks in large-batch llm inference.
\newblock {\em arXiv preprint arXiv:2503.08311}, 2025.

\bibitem{coqa}
Siva Reddy, Danqi Chen, and Christopher~D Manning.
\newblock Coqa: A conversational question answering challenge.
\newblock {\em Transactions of the Association for Computational Linguistics}, 2019.

\bibitem{winogrande}
Keisuke Sakaguchi, Ronan~Le Bras, Chandra Bhagavatula, and Yejin Choi.
\newblock Winogrande: An adversarial winograd schema challenge at scale.
\newblock {\em Communications of the ACM}, 2021.

\bibitem{layerbylayer}
Oscar Skean, Md~Rifat Arefin, Dan Zhao, et~al.
\newblock Layer by layer: Uncovering hidden representations in language models.
\newblock In {\em Forty-second International Conference on Machine Learning}, 2025.

\bibitem{painter}
Qi~Sun, Marc Pickett, Aakash~Kumar Nain, and Llion Jones.
\newblock Transformer layers as painters.
\newblock In {\em AAAI-25, Sponsored by the Association for the Advancement of Artificial Intelligence, February 25 - March 4, 2025, Philadelphia, PA, {USA}}, 2025.

\bibitem{llamaeagle}
Bangsheng Tang, Carl~Chengyan Fu, Fei Kou, Grigory Sizov, Haoci Zhang, Jason Park, Jiawen Liu, Jie You, Qirui Yang, Sachin Mehta, et~al.
\newblock Efficient speculative decoding for llama at scale: Challenges and solutions.
\newblock {\em arXiv preprint arXiv:2508.08192}, 2025.

\bibitem{gemma}
Gemma Team, Thomas Mesnard, Cassidy Hardin, Dadashi, et~al.
\newblock Gemma: Open models based on gemini research and technology.
\newblock {\em arXiv preprint arXiv:2403.08295}, 2024.

\bibitem{am}
Xiaoyu Tian, Yunjie Ji, Haotian Wang, Shuaiting Chen, Sitong Zhao, Yiping Peng, Han Zhao, and Xiangang Li.
\newblock Not all correct answers are equal: Why your distillation source matters.
\newblock {\em arXiv preprint arXiv:2505.14464}, 2025.

\bibitem{llama}
Hugo Touvron, Thibaut Lavril, Gautier Izacard, Xavier Martinet, and Lachaux...
\newblock Llama: Open and efficient foundation language models.
\newblock {\em arXiv preprint arXiv:2302.13971}, 2023.

\bibitem{transformer}
Ashish Vaswani, Noam Shazeer, Niki Parmar, et~al.
\newblock Attention is all you need.
\newblock In {\em Advances in Neural Information Processing Systems 30: Annual Conference on Neural Information Processing Systems 2017, December 4-9, 2017, Long Beach, CA, {USA}}, 2017.

\bibitem{middle}
Hanyu Wang, Bochuan Cao, Yuanpu Cao, and Jinghui Chen.
\newblock Truthflow: Truthful {LLM} generation via representation flow correction.
\newblock In {\em Forty-second International Conference on Machine Learning}, 2025.

\bibitem{valm}
Weizhi Wang, Li~Dong, Hao Cheng, Haoyu Song, Xiaodong Liu, Xifeng Yan, Jianfeng Gao, and Furu Wei.
\newblock Visually-augmented language modeling.
\newblock In {\em The Eleventh International Conference on Learning Representations}, 2023.

\bibitem{flan}
Jason Wei, Maarten Bosma, Vincent Zhao, Kelvin Guu, Adams~Wei Yu, Brian Lester, Nan Du, Andrew~M. Dai, and Quoc~V Le.
\newblock Finetuned language models are zero-shot learners.
\newblock In {\em International Conference on Learning Representations}, 2022.

\bibitem{cot}
Jason Wei, Xuezhi Wang, Dale Schuurmans, Maarten Bosma, Fei Xia, Ed~Chi, Quoc~V Le, Denny Zhou, et~al.
\newblock Chain-of-thought prompting elicits reasoning in large language models.
\newblock {\em Advances in neural information processing systems}, 2022.

\bibitem{qwen3}
An~Yang, Anfeng Li, Baosong Yang, Beichen Zhang, Binyuan Hui, Bo~Zheng, Bowen Yu, Chang Gao, Chengen Huang, Chenxu Lv, et~al.
\newblock Qwen3 technical report.
\newblock {\em arXiv preprint arXiv:2505.09388}, 2025.

\bibitem{edgemma}
Biao Zhang, Fedor Moiseev, Joshua Ainslie, Paul Suganthan, Min Ma, Surya Bhupatiraju, Fede Lebron, Orhan Firat, Armand Joulin, and Zhe Dong.
\newblock Encoder-decoder gemma: Improving the quality-efficiency trade-off via adaptation.
\newblock {\em arXiv preprint arXiv:2504.06225}, 2025.

\bibitem{late2}
Jianyi Zhang, Da-Cheng Juan, Cyrus Rashtchian, et~al.
\newblock {SLED}: Self logits evolution decoding for improving factuality in large language models.
\newblock In {\em The Thirty-eighth Annual Conference on Neural Information Processing Systems}, 2024.

\end{thebibliography}
\bibliographystyle{plain}
\end{document}